\documentclass[runningheads]{llncs}
\usepackage[T1]{fontenc}
\usepackage{graphicx}
\usepackage{amsmath}
\usepackage{amssymb}
\usepackage{bm} 
\usepackage{cleveref}

\usepackage{multirow}
\usepackage{pifont} 

\usepackage{xspace}
\usepackage{booktabs}
\usepackage{multirow}
\usepackage{pifont}     
\usepackage{siunitx}
\usepackage{adjustbox}  
\newcommand{\wimg}[1]{\raisebox{-0.1\height}{\includegraphics[height=8mm]{#1}}}

\newcommand{\KanjiGo}{\raisebox{-.2\height}{\includegraphics[width=1em]{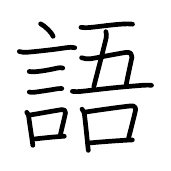}}}
\newcommand{\KanjiGen}{\raisebox{-.2\height}{\includegraphics[width=1em]{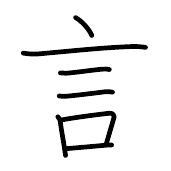}}}
\newcommand{\KanjiKou}{\raisebox{-.2\height}{\includegraphics[width=1em]{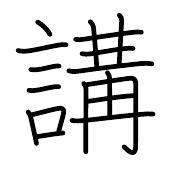}}}
\newcommand{\KanjiKei}{\raisebox{-.2\height}{\includegraphics[width=1em]{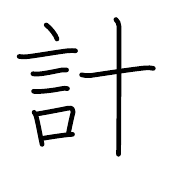}}}
\newcommand{\KanjiYoku}{\raisebox{-.2\height}{\includegraphics[width=1em]{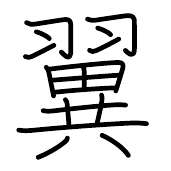}}}

\crefname{figure}{Fig.}{Figs.}
\crefname{table}{Table}{Tables}
\crefname{section}{Sec.}{Secs.}
\crefname{equation}{Eq.}{Eqs.}

\begin{document}
\title{Handwriting Trajectory Recovery with \\ Diffusion Models}
%
%


\author{
Hiroki Nagamatsu\orcidID{0009-0005-2038-4923} 
\and
Shoji Toyota\orcidID{0000-0003-0613-4205}
\and
Seiichi Uchida\orcidID{0000-0001-8592-7566}
}
%
\authorrunning{H. Nagamatsu et al.}
%
\institute{Kyushu University, Fukuoka, Japan\\
\email{hiroki.nagamatsu@human.ait.kyushu-u.ac.jp}}
\maketitle              
\begin{abstract}
Recovering online pen trajectories from offline handwriting images, often referred to as handwriting trajectory recovery (stroke recovery), is an offline-to-online conversion task with applications in stroke-level editing and forensic analysis.
We propose, to the best of our knowledge, the first diffusion-model-based framework for this task. Our method formulates trajectory recovery as image-conditioned generation and uses a denoising diffusion model to sample pen trajectories consistent with the observed ink trace.
Through extensive quantitative evaluations on CASIA-OLHWDB (1.0--1.1), we verify that the proposed approach enables accurate recovery even for complex multi-stroke characters, substantially improving both temporal similarity (DTW/LDTW) and shape fidelity (AIoU) over representative prior methods such as PEN-Net and Cross-VAE.
We further show that the model captures general stroke-order tendencies and generalizes to classes unseen during training, exemplified by cross-script transfer: a model trained on Chinese characters can recover reasonable stroke orders for Latin letters to some extent.

\end{abstract}

\section{Introduction}
\label{sec:intro}
\emph{Handwriting trajectory recovery} (also referred to as \emph{stroke recovery}) is an \emph{offline-to-online conversion} task: given a completed offline handwriting image, the goal is to reconstruct a time-ordered pen trajectory that is consistent with the observed ink trace.
Recovering the trajectory turns a static raster image into a dynamic ``digital ink'' representation, enabling stroke-level operations (e.g., retouching or deleting individual strokes) and resolution-independent rendering by resampling the continuous path before rasterization.
Temporal cues are also valuable for downstream analysis --- for example, directionality and pen dynamics can improve comparisons in signature verification and related tasks~\cite{Survey}.

\begin{figure}[t]
  \centering
  \includegraphics[width=0.9\textwidth]{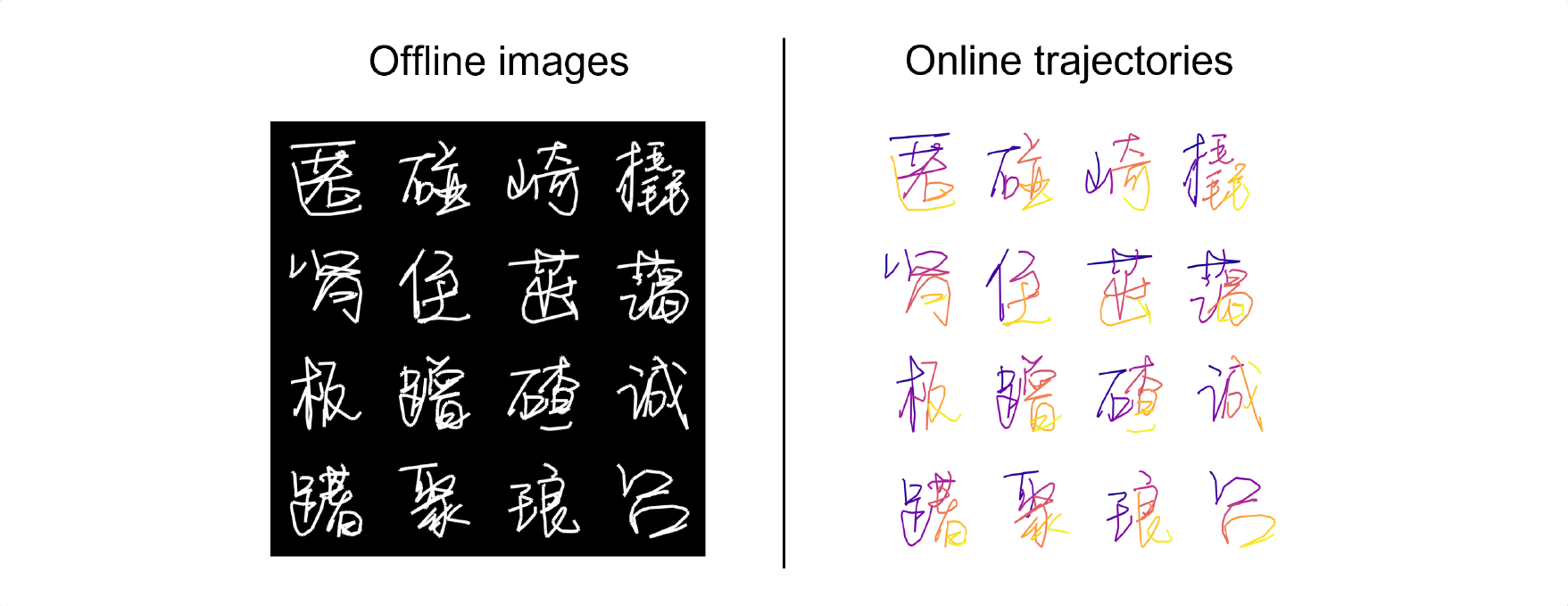}\\[-2mm]
  \caption{Example of an offline character image and its corresponding online handwriting trajectory.}
  \label{fig:offline_online_example}
\end{figure}

Fig.~\ref{fig:offline_online_example} illustrates the modality gap between an offline character image and its underlying online trajectory.
While rasterizing a trajectory into an image (often called \emph{inking}) is straightforward, the inverse mapping is fundamentally ill-posed: a static image records only the final ink trace, not the writing process that produced it.
For simple glyphs (e.g., letters or digits with few strokes and limited pen-up events), heuristic rules can sometimes recover a plausible order~\cite{Rule_base_old,Rule_base_old2}, but ambiguity grows rapidly for complex multi-stroke characters where crossings, tight loops, and stroke concatenations are frequent.

The goal of this paper is to experimentally demonstrate that the handwriting trajectory recovery task for complex multi-stroke characters can be solved with high accuracy using {\em diffusion models}~\cite{diffusion,DDPM}. Although diffusion models are widely used for image synthesis, the same denoising framework can also generate 1D time-series data by using a 1D denoiser.
For instance, a model trained on many samples of the 14-stroke character ``\KanjiGo'' can stochastically generate diverse trajectories of ``\KanjiGo'' by changing the initial noise.

In trajectory recovery, however, the objective is not to generate diverse trajectories arbitrarily, but to generate the particular trajectory that corresponds to the observed ink trace in a handwriting image.
To this end, we propose an image-conditioned diffusion model for time-series generation, where the input handwriting image guides the denoising process. Moreover, to obtain more accurate estimates, we introduce multi-scale conditioning, which gradually injects image evidence from coarse global shape down to fine local details. With this model, even very complex multi-stroke characters can be recovered with high precision.

Several learning-based models for handwriting trajectory recovery have been proposed, including PEN-Net~\cite{PEN-Net} and Cross-VAE~\cite{Cross-VAE}.
However, they often struggle with the complexity of multi-stroke characters, and their recovered trajectories may fail to preserve the original glyph shape.
In particular, the predicted pen path can drift away from the observed ink trace, producing pen trajectories that pass through regions where no stroke exists in the input image.
In contrast, by leveraging diffusion models for the trajectory recovery task --- to the best of our knowledge, for the first time --- we aim to address these limitations.

Experiments demonstrate that the proposed model achieves accurate recovery even for complex characters such as those in Fig.~\ref{fig:offline_online_example}.
Interestingly, we also observe that recovery accuracy improves as the number of training character classes increases (e.g., from 20 classes to 3,000 classes). This suggests that handwriting trajectories have general trends across character classes. One plausible source of this regularity is the reuse of radicals across characters. Although the characters ``\KanjiGo,'' ``\KanjiKei,'' and ``\KanjiKou'' are different classes, they share the radical ``\KanjiGen,'' and their handwriting trajectories are largely consistent; therefore, more classes and more training data can directly improve recovery accuracy.
Beyond such radical-level trends, our experiments indicate that the model also learns more general writing tendencies (e.g., left-to-right and top-to-bottom tendencies).
Concretely, a model trained on Chinese characters can recover stroke orders for Latin letters to some extent, demonstrating cross-script generalization.

Our main contributions are as follows.
\begin{itemize}
  \item We address handwriting trajectory recovery with diffusion models for the first time. Specifically, we propose an image-conditioned diffusion model for time-series generation, and show that multi-resolution conditioning further improves recovery accuracy.
  \item Experiments on CASIA-OLHWDB show clear improvements over PEN-Net and Cross-VAE in terms of both temporal alignment (DTW/LDTW) and shape fidelity (AIoU).
  \item We experimentally demonstrate that the learned model generalizes to not only unseen Chinese character classes but also unseen Latin alphabet characters. We also show that it produces plausible recovery results even for ``incomplete'' characters where parts of strokes are intentionally removed.
\end{itemize}

\section{Related Work}
As reviewed in~\cite{Survey}, classical handwriting trajectory recovery methods rely on heuristic rules to resolve ambiguous structures such as stroke intersections, overlapping strokes (double-strokes), and loops. In general, these methods (i)~detect keypoints such as intersections and endpoints, (ii)~decompose the ink trace into sub-strokes based on these keypoints, and 
(iii)~reconnect the sub-strokes according to hand-crafted rules. For instance, a ``+''-shaped intersection may be split into four sub-strokes and then reconnected into a horizontal segment ``--'' and a vertical segment ``|,'' rather than two corner-like fragments (e.g., ``$\llcorner$'' and ``$\urcorner$''). Such rule-based methods have mainly been applied to single-stroke characters (or single-stroke cursive words)~\cite{Rule_base,Rule_base_old2,Rule_base2,Rule_base_old}, and have rarely been used for more complex multi-stroke characters, because multi-stroke characters introduce many more intersections and ambiguities, making rule-based reconnection impractical.\par

Deep neural networks have increasingly been used for handwriting trajectory recovery, with different design choices on how strongly the output is constrained by intermediate structures. One line of work explicitly decomposes the input into sub-strokes and then predicts their temporal evolution or ordering, like the above classical methods. Pen Tip Motion Prediction (PTMP) ~\cite{PEN_Tip} tackles drawing-order recovery through pen-tip motion prediction: a CNN takes partially drawn handwriting images as input and outputs a distribution over the next pen position, which is then used in an iterative framework to generate the pen movement sequence. Setsort~\cite{Setsort} further constrains the solution space at the sub-stroke level by first inferring a skeleton with a CNN and cutting it into sub-strokes; a Sub-stroke Encoding Transformer (SET) embeds each sub-stroke point sequence, and a Sub-strokes Ordering Transformer (SORT) predicts the discrete ordering together with the pen state. Such structure-constrained pipelines can produce geometrically precise reconstructions when the extracted skeleton/sub-strokes are reliable, but they inherit the limitations of classical methods on densely written multi-stroke characters.

Another line of work learns a more direct mapping from an offline image to an online coordinate sequence using encoder--decoder architectures, reducing reliance on explicit stroke extraction. DED-Net~\cite{DED-Net} proposes an end-to-end deep encoder--decoder network that encodes the offline image into features and decodes them into a trajectory point sequence.
Kanji-Net~\cite{Kanji-Net} adapts encoder--decoder recovery to multi-stroke characters by incorporating an attention mechanism to focus on local context during sequential generation and by using a mixture density output (GMM) to improve robustness to handwriting distortions and writer variation. PEN-Net~\cite{PEN-Net} specifically targets complex glyphs and long trajectories with a parsing-and-tracing encoder--decoder: it introduces a double-stream parsing encoder to capture complementary horizontal/vertical context, and a global tracing decoder that conditions each decoding step on glyph parsing features to mitigate trajectory-point drifting; it also proposes AIoU and LDTW to jointly evaluate glyph fidelity and temporal alignment for complex characters.
From a different perspective, Cross-VAE~\cite{Cross-VAE} performs bidirectional modality conversion by coupling an image VAE and a trajectory VAE through a shared latent space with an explicit space-sharing loss, enabling stroke recovery by encoding the offline image and decoding it into an online trajectory. Despite these advances, recovering accurate stroke order while maintaining strict geometric consistency with the input ink trace remains challenging for highly complex multi-stroke characters, and several methods still exhibit noticeable drift or inconsistencies in difficult regions.

In contrast to the above pipelines, we introduce diffusion models to handwriting trajectory recovery for the first time, to the best of our knowledge. Our approach performs image-conditioned denoising directly on a 1D trajectory sequence: starting from Gaussian noise, a 1D denoiser iteratively updates the \emph{entire} trajectory while being guided by the input handwriting image through multi-resolution conditioning~\cite{diffusion,DDPM}. This design mitigates two recurring limitations of prior methods.
First, it avoids fragile intermediate representations such as explicit skeletonization and sub-stroke segmentation by letting the model refer to image evidence throughout the iterative refinement process. Second, unlike single-pass or strictly step-by-step decoding, iterative denoising enables global corrections to the trajectory over multiple refinement steps, which helps suppress geometric drift from the observed ink trace, especially for long and complex multi-stroke characters. As demonstrated in our experiments, the resulting model consistently surpasses existing approaches in both temporal alignment and glyph fidelity on complex characters.

\section{Handwriting Trajectory Recovery with Diffusion Models}
\subsection{Problem Formulation}
\begin{figure}[t]
  \centering
  \includegraphics[width=\textwidth]{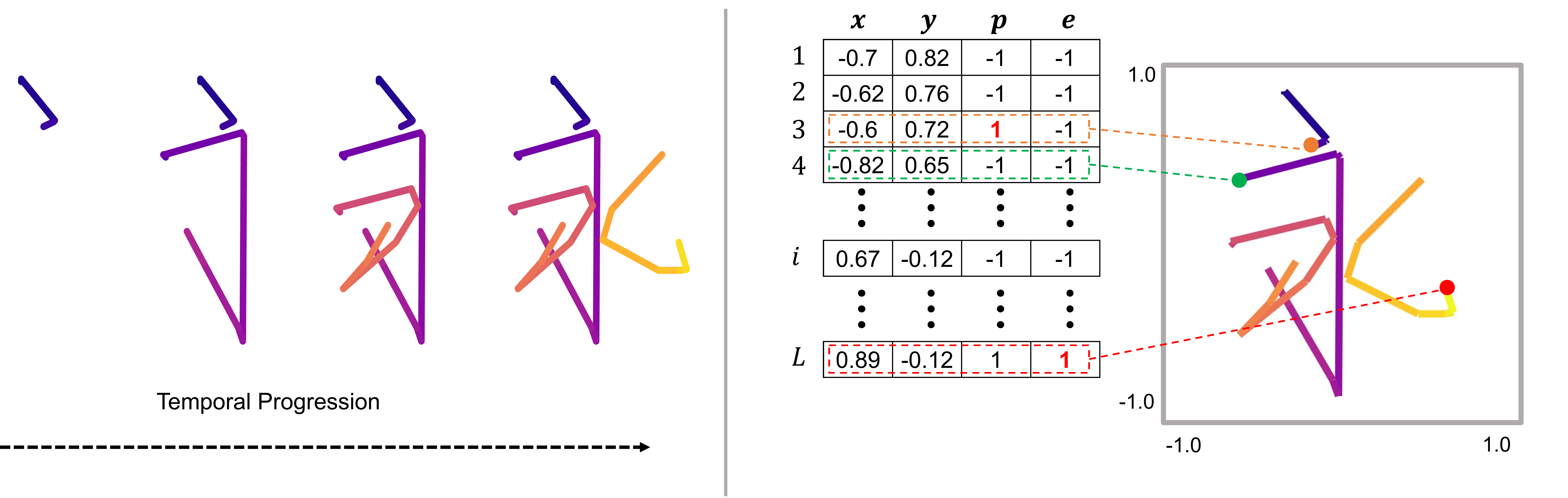}\\[-2mm]
  \caption{Visualization of an online trajectory sequence. The color changes from dark to bright along the temporal order.}
  \label{fig:trajectory_visualization}
\end{figure}

We aim to sample plausible online handwriting trajectories conditioned on an offline image. 
Let $I \in \mathbb{R}^{H\times W}$ be an offline handwriting image with height $H \in \mathbb{N}_{>0}$ and width $W \in \mathbb{N}_{>0}$, and let $\mathbf{X}$ be an online trajectory sequence:
\begin{equation}
\mathbf{X}=\{(x_i,y_i,p_i,e_i)\}_{i=1}^{L}.
\end{equation}
Fig.~\ref{fig:trajectory_visualization} illustrates $\mathbf{X}$. Here $(x_i, y_i)$ denotes the pen coordinate at sequence index $i$.
$p_i \in \{-1,1\}$ denotes a binary variable indicating whether the pen is up ($1$) or down ($-1$). 
$e_i \in \{-1,1\}$ is an end-of-sequence (EOS) flag. Once $e_i=1$ at the terminal point, we pad the remaining positions up to the maximum length $L$ by repeating the terminal coordinate.
\par

Stroke recovery from an offline image is an ill-posed inverse problem, because multiple plausible writing trajectories can produce visually similar (or even identical) rasterized character images.
Therefore, instead of learning a deterministic image-to-sequence mapping, we formulate the task as learning a conditional distribution over trajectories, $p(\mathbf{X}\mid I)$.

For stable learning, we normalize the coordinates $(x_i, y_i)$ to $[-1,1]$. 
Although $p_i$ and $e_i$ are inherently discrete variables, we treat them as continuous values
in the diffusion model and binarize them using a threshold of 0 at the final output stage.
Let $\mathcal{D}=\{(I^{(n)}, \mathbf{X}^{(n)})\}_{n=1}^{N}$ be a set of paired training samples.
Our goal is to learn the conditional distribution $p_{\theta}(\mathbf{X}\mid I)$ and recover trajectories by sampling from it.

\subsection{Conditional Diffusion Model for Trajectories}
A diffusion model \cite{diffusion,DDPM} is a powerful approach for generating samples from a target distribution. While diffusion models are typically used to generate images or videos, they can also be applied to the generation of handwriting trajectories. 

In the diffusion model framework, we define a forward diffusion process 
$\mathbf{X}_0 \rightarrow \mathbf{X}_1 \rightarrow \cdots \rightarrow \mathbf{X}_T$ 
over handwriting trajectories by gradually adding Gaussian noise to a clean handwriting trajectory $\mathbf{X}_0=\mathbf{X}$ over $T$ diffusion steps. 
Here, $t \in \{1, \dots, T\}$ denotes the diffusion step, and $\mathbf{X}_t$ represents the noisy handwriting trajectory at step $t$. 
Given a variance schedule $\{\beta_t\}_{t=1}^{T}$ and $\alpha_t=1-\beta_t$, the noised handwriting trajectory at step $t$ is expressed as
\begin{equation}\label{eq:diffusion_forward}
\mathbf{X}_t=\sqrt{\bar{\alpha}_t}\,\mathbf{X}_0+\sqrt{1-\bar{\alpha}_t}\,\boldsymbol{\epsilon},\quad
\boldsymbol{\epsilon}\sim\mathcal{N}(\mathbf{0},\mathbf{I}),
\end{equation}
where $\bar{\alpha}_t=\prod_{s=1}^{t}\alpha_s$.

By considering the reverse process 
$\mathbf{X}_T \rightarrow \mathbf{X}_{T-1} \rightarrow \cdots \rightarrow \mathbf{X}_0$ 
corresponding to \eqref{eq:diffusion_forward}, we can generate a handwriting trajectory from the target distribution of $\mathbf{X}_0$. 
To model the reverse process, it is sufficient to estimate the noise term $\boldsymbol{\epsilon}(\mathbf{X}_t, t)$, which can be learned using a denoising network $\boldsymbol{\epsilon}_{\theta}(\mathbf{X}_t, t)$~\cite{DDPM}.

While the diffusion models described above enable the sampling of random handwriting trajectories, 
our objective is to generate a trajectory that is aligned with a given offline image $I$, rather than an arbitrary random sample. 
To this end, we incorporate $I$ as a conditioning variable into the denoising network. Formally, we define the conditioned denoising network $\epsilon_{\theta}$ as
\begin{equation}
\hat{\boldsymbol{\epsilon}}=\epsilon_{\theta}(\mathbf{X}_t,t,f_{\phi}(I)),
\end{equation}
where $f_{\phi}$ denotes an image encoder applied to $I$. 
Using this conditioned denoising network, a handwriting trajectory is generated by iteratively denoising a sample drawn from a Gaussian distribution, with the generation process conditioned on $I$. 

To train the denoising network, we adopt the standard noise-prediction objective:
\begin{equation}
\mathcal{L}(\theta, \phi)=\mathbb{E}_{t, I, \mathbf{X}_0,\boldsymbol{\epsilon}}
\left[
\left\|
\boldsymbol{\epsilon}-\epsilon_{\theta}(\mathbf{X}_t,t,f_{\phi}(I))
\right\|_2^2
\right].
\end{equation}
The expectation $\mathbb{E}_{t, I, \mathbf{X}_0,\boldsymbol{\epsilon}}$ can be empirically estimated using the training dataset 
$\mathcal{D}=\{(I^{(n)}, \mathbf{X}^{(n)})\}_{n=1}^{N}$ consisting of offline images and their corresponding trajectories, 
where the time index $t$ is sampled uniformly from $\{1,\dots,T\}$ for each training example.

\subsection{Multi-scale Image Conditioning}
\label{sec:archi}

\begin{figure}[t]
  \centering
  \includegraphics[width=0.95\textwidth]{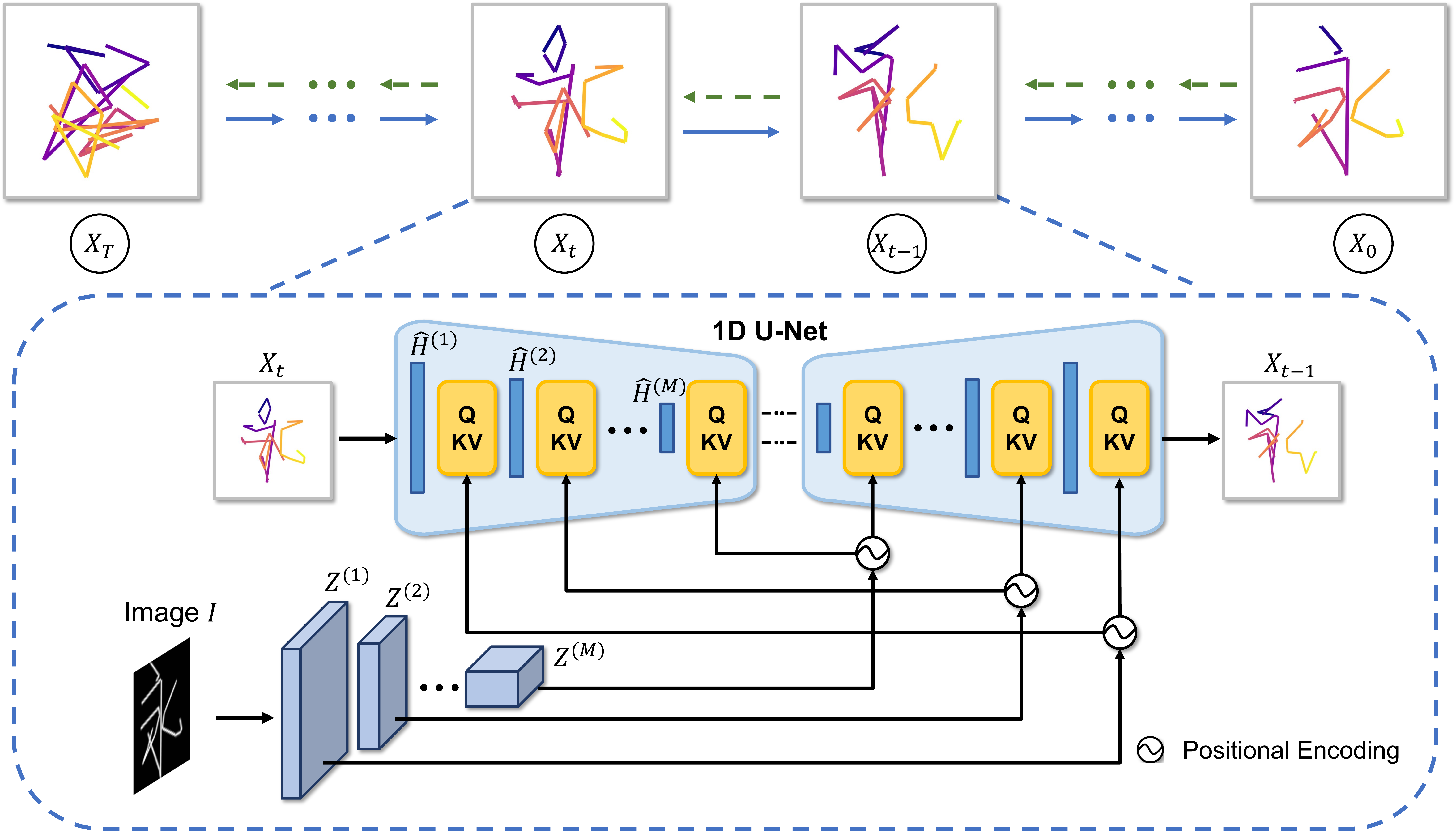}\\[-2mm]
  \caption{Overview of the proposed image conditioned diffusion model for handwriting trajectory recovery.}
  \label{fig:overview}
\end{figure}

Accurate stroke recovery requires preserving the spatial structure of the input image, because the offline character image provides strong geometric constraints on plausible pen trajectories.
Therefore, instead of compressing the image into a single global latent vector, we condition the denoiser on multi-resolution image feature maps that keep spatially localized features.
This is particularly important in ambiguous regions (e.g., crossings, closely adjacent strokes, and loops), where local geometric evidence must be aligned with trajectory tokens.

An overview of the full architecture of the proposed model is shown in Fig.~\ref{fig:overview}.
Let the image encoder $f_{\phi}$ produce $M$ feature maps at different resolutions:
\begin{equation}
\mathcal{Z} = f_{\phi}(I)=\{Z^{(m)}\}_{m=1}^{M}, \qquad
Z^{(m)} \in \mathbb{R}^{H_m \times W_m \times C_m},
\end{equation}
where $m$ indexes the feature scale, and $H_m$, $W_m$, and $C_m$ denote the height, width, and channel dimensions of the $m$-th feature map, respectively.
For each scale $m$, we project the feature map to the denoiser hidden dimension $d$, add a 2D positional encoding, and flatten it into spatial tokens:
\begin{equation}
\tilde{Z}^{(m)} = \mathrm{Flatten}\!\left(g^{(m)}(Z^{(m)}) + E^{(m)}_{\mathrm{2D}}\right)
\in \mathbb{R}^{N_m \times d},
\qquad N_m = H_m W_m,
\end{equation}
where $g^{(m)}(\cdot)$ denotes a scale-specific projection layer used to align the channel dimension to $d$, and $E^{(m)}_{\mathrm{2D}}$ denotes the 2D positional encoding for the $m$-th feature map (with the same spatial resolution as $Z^{(m)}$).
$\mathrm{Flatten}(\cdot)$ reshapes the 2D feature grid into a sequence of $N_m$ spatial tokens.

Our denoiser is a 1D U-Net~\cite{unet} operating along the trajectory sequence axis, with 1D convolutional blocks for local continuity and self-attention layers for long-range dependencies.
To inject image evidence, we insert cross-attention blocks at selected U-Net stages.
At each U-Net resolution, we inject the corresponding image-feature scale via cross-attention.
Let $\hat{H}^{(m)} \in \mathbb{R}^{L_m \times d}$ denote the trajectory features at the U-Net stage paired with image scale $m$, where $L_m$ is the trajectory token length at that stage.
We then apply cross-attention with trajectory tokens as queries and image tokens as keys and values:
\begin{equation}
H'^{(m)}
=
\hat{H}^{(m)}
+
\mathrm{MHCA}\!\left(
\hat{H}^{(m)}W_Q^{(m)},\ \tilde{Z}^{(m)}W_K^{(m)},\ \tilde{Z}^{(m)}W_V^{(m)}
\right),
\end{equation}
where $\mathrm{MHCA}$ denotes standard multi-head cross-attention, and $W_Q^{(m)}$, $W_K^{(m)}$, and $W_V^{(m)}$ are learnable linear projection matrices for queries, keys, and values, respectively.

This design preserves explicit spatial correspondence between image regions and trajectory tokens, unlike conditioning with a single global vector.
Moreover, multi-scale feature injection provides complementary cues: coarse-scale features capture the global character layout, while fine-scale features preserve local stroke geometry.
The effect of this design is examined in the ablation study in Section~\ref{sec:injection}.

\section{Experiments}
\subsection{Experimental Setup}

\subsubsection{Chinese Handwriting Dataset}

We evaluate the performance of our method on large-scale Chinese online handwriting datasets, CASIA-OLHWDB 1.0 and 1.1~\cite{casia}, which contain samples from 420 and 300 writers, respectively, totaling 720 writers.
In this study, we focus on the 3,755 character classes in the GB2312-80 Level-1 set.
We create writer-disjoint splits with 576 writers for training and 144 writers for testing.
All reported results are computed on samples written by the 144 test writers.
To evaluate generalization to unseen character classes, we split the 3,755 classes into 3,000 seen classes and 755 unseen classes.
We train the model using only samples from the seen classes.
We then report performance on the 3,000 seen classes and the 755 unseen classes.

For fair comparison with prior work, we follow the same offline-image rendering process as PEN-Net \cite{PEN-Net}.
Offline images are rendered from online trajectories.
For each trajectory, we rasterize the ink trace into a $64\times64$ image with a black background and white strokes.

\subsubsection{Implementation Details}
\label{sec:imple}
We use $T{=}1000$ diffusion steps with a cosine noise schedule~\cite{Cos_schedule}.
During testing, we use the DDPM sampler \cite{DDPM} with 1,000 steps.
We optimize the model using Adam \cite{kingma2014adam} with a batch size of 512, a learning rate of $8\times10^{-5}$, and 500{,}000 training steps.
The experiments are conducted on a single A100 GPU.

\subsubsection{Evaluation Metrics}

We evaluate recovery quality from two complementary viewpoints: temporal correctness and spatial glyph fidelity.
For temporal correctness, we use Dynamic Time Warping (DTW) and Length-independent DTW (LDTW)~\cite{PEN-Net} as trajectory distances.
DTW aligns two sequences by an optimal warping path and accumulates point-wise Euclidean distances; lower values indicate better temporal similarity.
Because DTW increases with sequence length, we also use LDTW.
LDTW normalizes the accumulated DTW cost by the warping-path length, enabling fair comparison across characters with different stroke counts.
For DTW/LDTW computation, we rescale coordinates to the image coordinate space $[0,64]$, so LDTW can be interpreted as an average pixel-level deviation from the ground-truth trajectory.
In our evaluation, the ground-truth stroke order and direction are defined by the temporal sequence recorded in the online CASIA data, rather than by a canonical dictionary order.

For spatial glyph fidelity, we use Adaptive IoU (AIoU)~\cite{PEN-Net}, which measures the overlap between (i) the ink region obtained by rasterizing the recovered trajectory and (ii) the ground-truth ink region.
Unlike standard IoU, AIoU adaptively adjusts the rasterization stroke width, making the metric less sensitive to minor stroke-thickness differences and better reflecting pure glyph-shape fidelity~\cite{PEN-Net}.

\begin{table}[t]
\caption{Trajectory recovery performance on seen classes (trained on 3,000 classes). Lower is better for DTW/LDTW; higher is better for AIoU.}
\label{tab:seen}
\centering
\begin{tabular}{lccc}
\toprule
Method & DTW$\downarrow$ & LDTW$\downarrow$ & AIoU$\uparrow$\\
\midrule
Cross-VAE~\cite{Cross-VAE} & 713.8 & 7.029 & 0.346\\
PEN-Net~\cite{PEN-Net}    & \underline{212.3} & \underline{2.728} & \underline{0.607}\\
Ours        & \textbf{111.1} & \textbf{1.574} & \textbf{0.745}\\
\bottomrule
\end{tabular}
\end{table}

\subsubsection{Baselines}

We compare our method with two representative deep learning baselines for handwriting trajectory recovery.
We use PEN-Net~\cite{PEN-Net} as a primary baseline because it is one of the most directly comparable methods for complex character trajectory recovery under similar datasets and evaluation metrics.
We additionally include Cross-VAE~\cite{Cross-VAE} as a complementary baseline; this is because Cross-VAE uses a non-autoregressive decoder, similarly to our approach.
Since their official implementations are not publicly available, we re-implement both baselines following the original papers and tune hyperparameters to reproduce the reported behavior.
\subsection{Main Results on Seen Classes}

\begin{figure}[t]
  \centering
  \includegraphics[width=\textwidth]{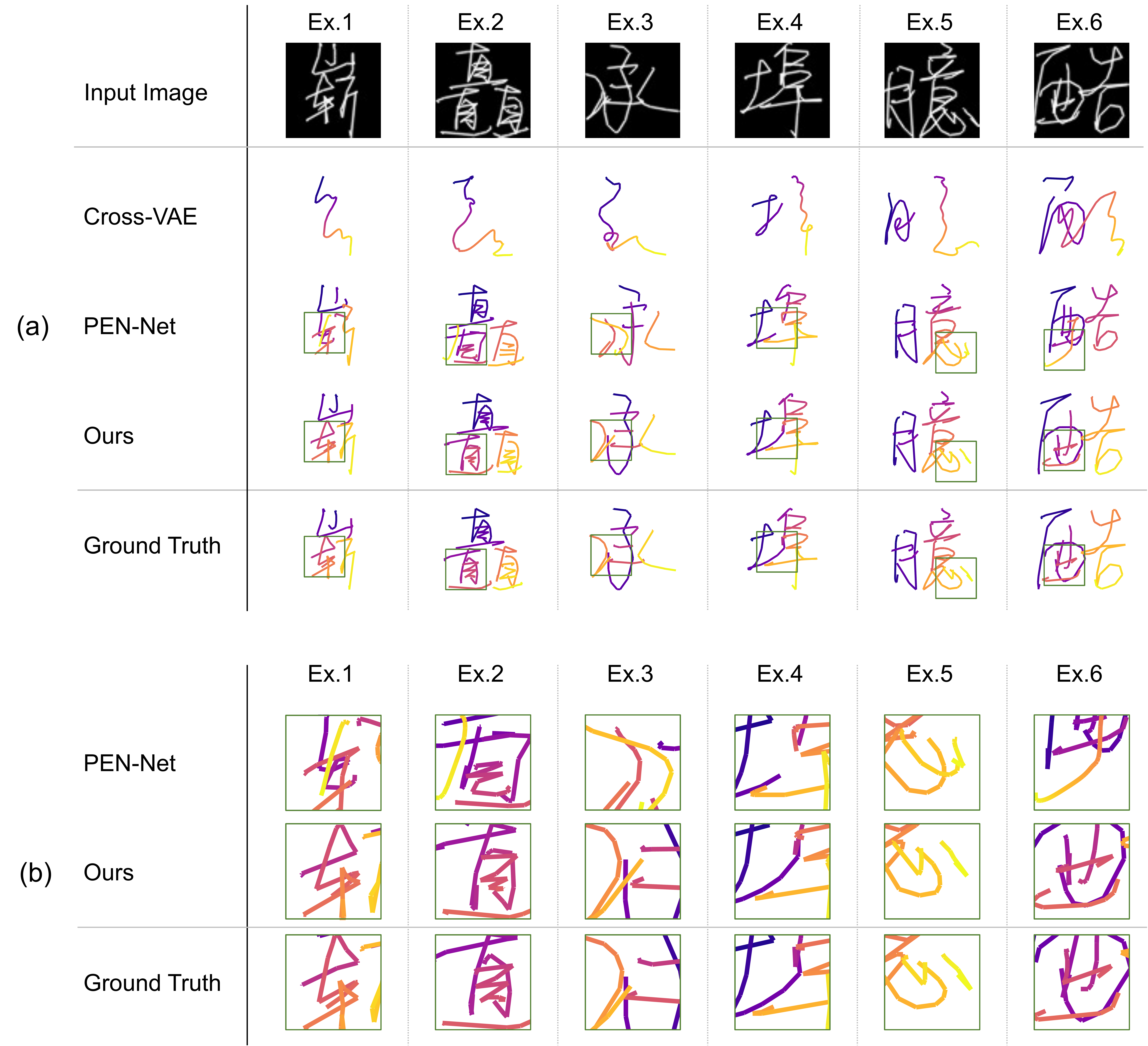}\\[-2mm]
  \caption{Qualitative comparison between our method and baselines in the 3,000-character-class setting. (a) Overall trajectory comparisons for representative examples. (b) Zoomed-in views of the green rectangular regions in (a), highlighting local differences in geometric fidelity.}
  \label{fig:qualitative_3000}
\end{figure}

We first report the main benchmark results on the seen classes test set.
Table~\ref{tab:seen} shows that our method achieves the best performance across all metrics.
In handwriting trajectory recovery, the offline image specifies the target ink layout; 
altering the glyph shape is invalid because it no longer corresponds to the given input and effectively becomes trajectory synthesis for a different shape.
Therefore, evaluation should prioritize fidelity to the given rasterized shape while assessing ordering consistency.
DTW/LDTW are affected by both temporal alignment and geometric deviations, so they do not isolate stroke order on their own. In particular, geometric drift can increase DTW/LDTW even when the stroke order is correct, whereas improved geometry can reduce them without fully resolving ordering errors.

Qualitative comparisons further show that our method preserves fine-grained stroke geometry while maintaining a plausible stroke order.
Fig.~\ref{fig:qualitative_3000} compares recovered trajectories from our method and the baselines on representative test samples.
Our method reproduces fine-grained details such as terminal hooks and curvature changes at turning points, and the color-coded temporal progression aligns well with the ground truth, suggesting accurate stroke-order recovery.
In contrast, PEN-Net occasionally produces spurious strokes or local 
mistakes that reduce 
glyph-shape consistency, and Cross-VAE often fails to reconstruct stroke-level structure for complex characters.
\begin{figure}[t]
  \centering
  \includegraphics[width=\textwidth]{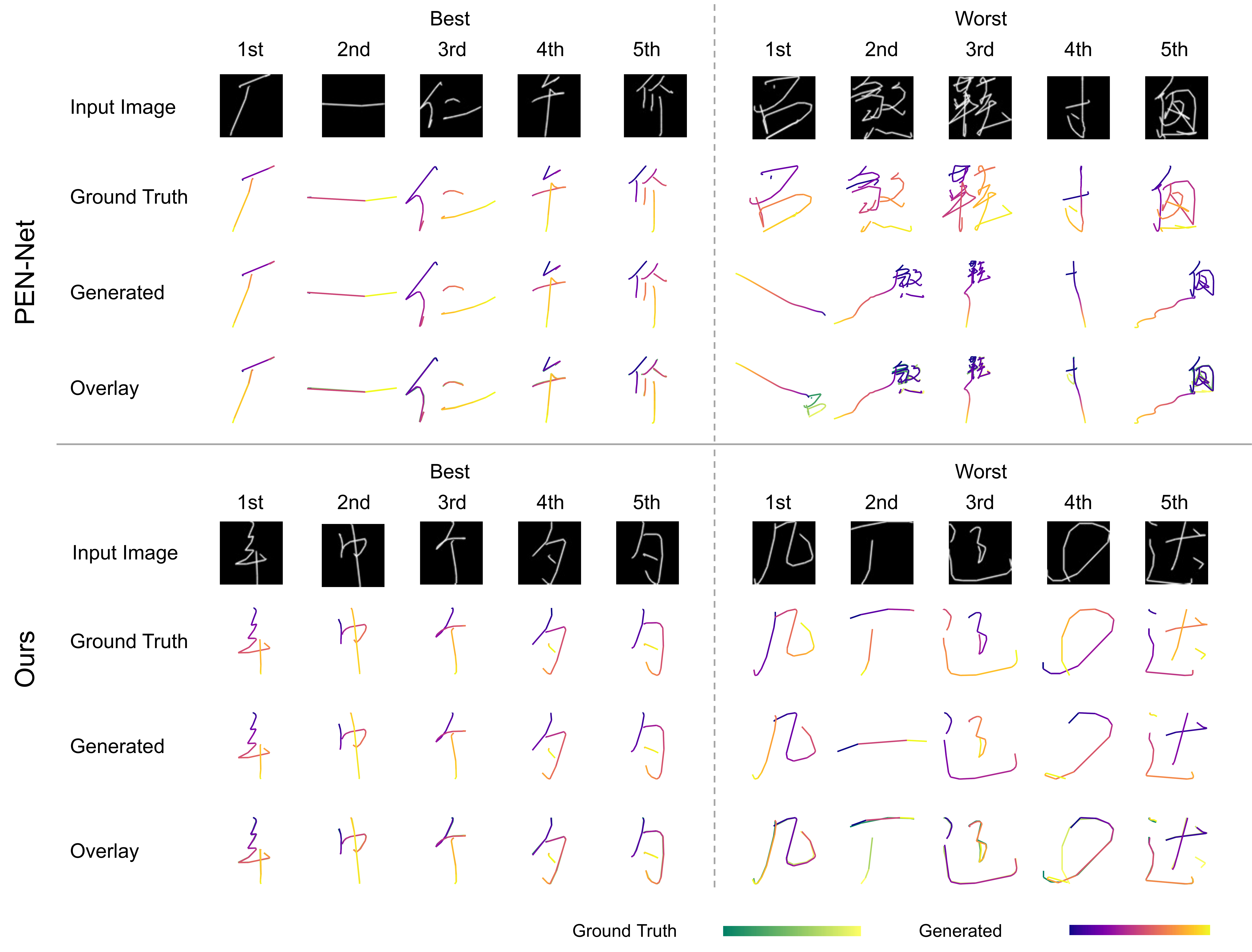}\\[-2mm]
  \caption{Five best and five worst qualitative examples ranked by LDTW for our method and PEN-Net (trained on 3,000 classes).}
  \label{fig:top_worst_ours_pennet}
\end{figure}

\subsection{Error Analysis: Ours vs. PEN-Net}

\begin{figure}[t]
  \centering
  \includegraphics[width=0.7\textwidth]{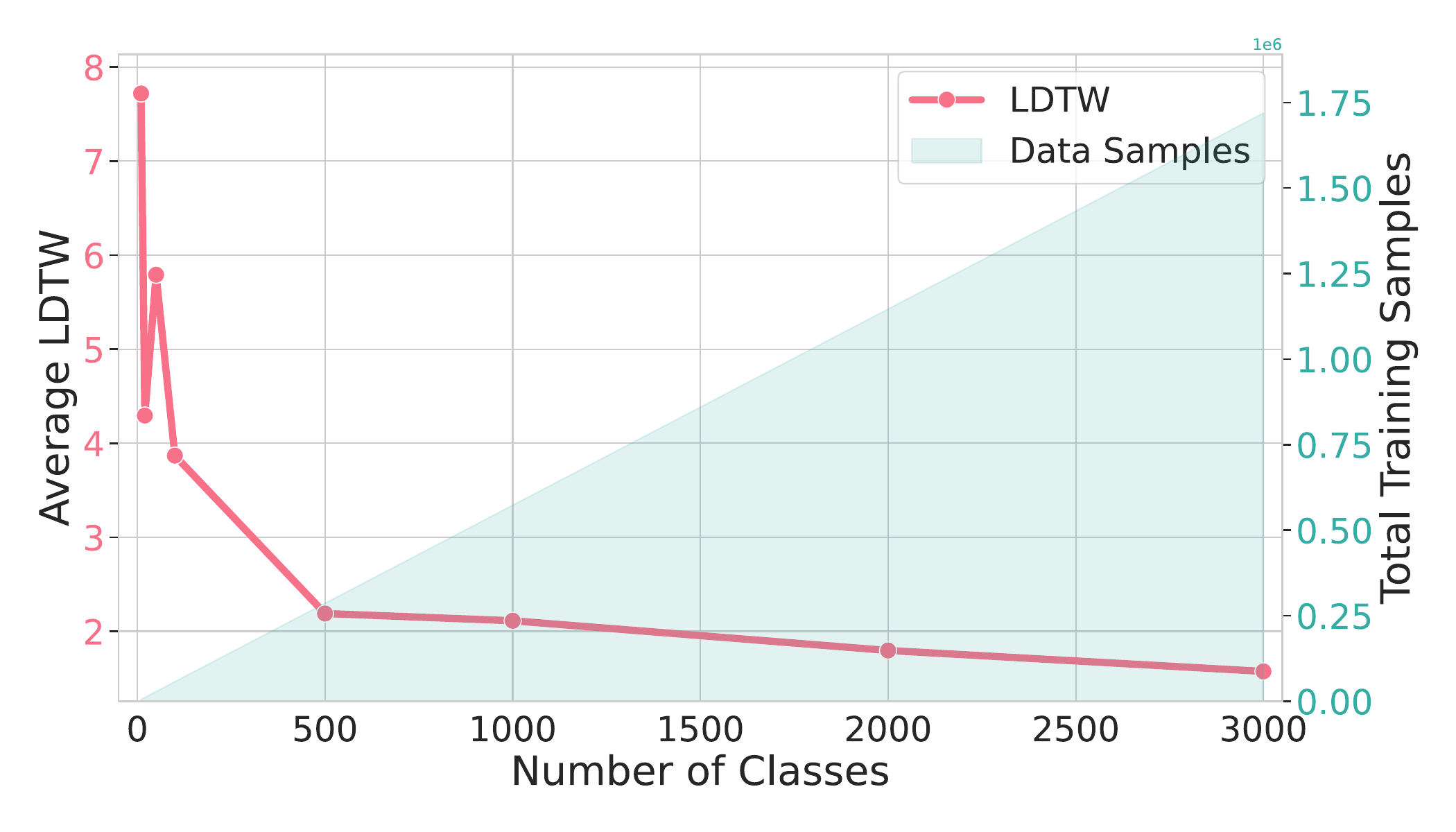}\\[-2mm]
  \caption{Test-set LDTW as the number of training character classes increases. The left y-axis shows LDTW, and the right y-axis shows the number of training samples (in $10^6$).}
  \label{fig:ldtw_vs_classes}
\end{figure}

\begin{figure}[t]
  \centering
  \includegraphics[width=0.9\textwidth]{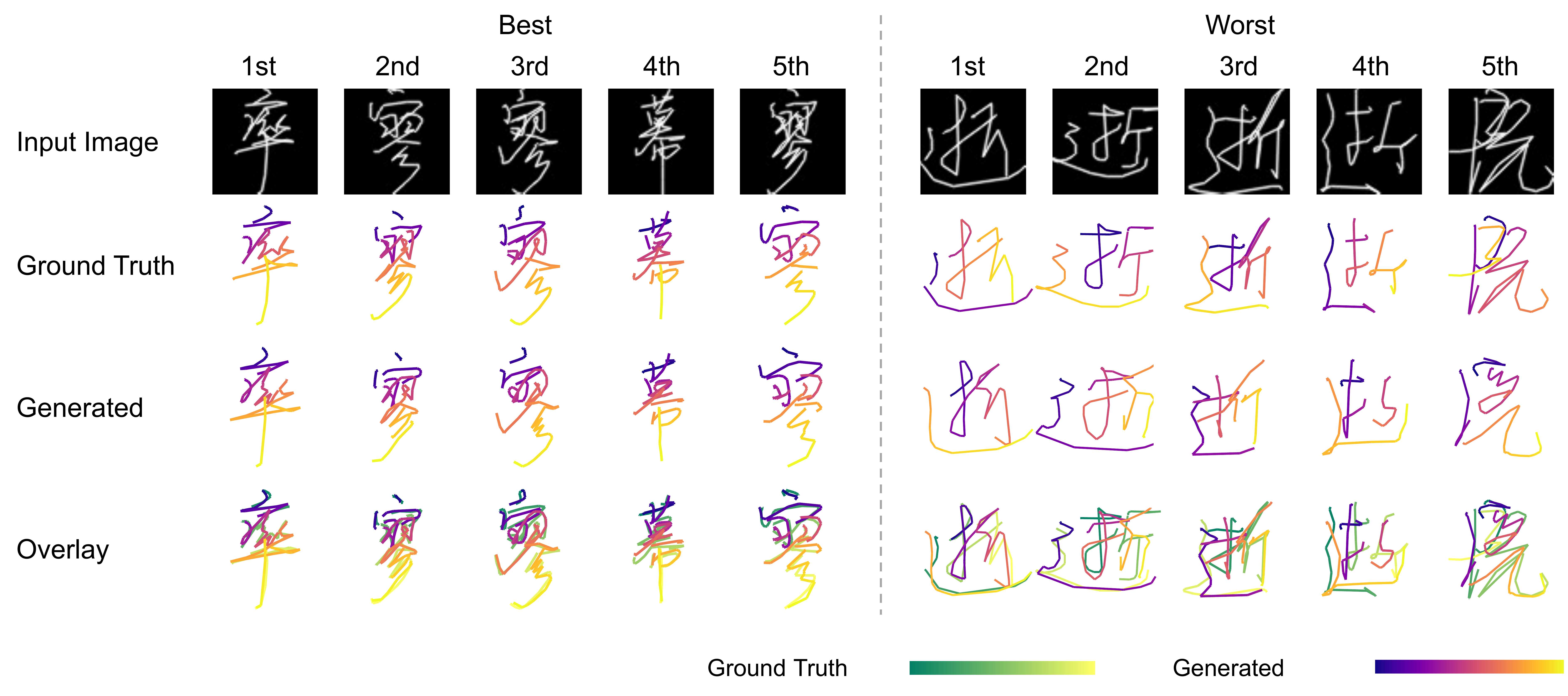}\\[-2mm]
  \caption{Qualitative results when training on 20 character classes. We show the five best and five worst ranked by LDTW.}
  \label{fig:qualitative_20}
\end{figure}

A major difference between our diffusion-model-based method and conventional approaches is that the latter are typically autoregressive: they recursively predict the next pen stroke based solely on the previous ones. In contrast, diffusion-model-based approaches generate pen strokes while incorporating information from both past and future time steps. As a result, in autoregressive methods, once an error occurs in the trajectory, it accumulates progressively over future time steps. By contrast, diffusion models are expected to alleviate such error accumulation, leading to superior performance compared to conventional methods.

To investigate this conjecture, we visualize the five best and five worst test cases ranked by LDTW for both our method and the autoregressive baseline, PEN-Net, as shown in Fig.~\ref{fig:top_worst_ours_pennet}.
Both methods perform well on relatively simple characters in the five best examples.
However, in the five worst cases, PEN-Net frequently exhibits unnaturally elongated strokes and large local deviations from the ink trace.
Because PEN-Net predicts coordinates sequentially conditioned on previous predictions, local mistakes can propagate and amplify over time, eventually causing severe trajectory breakdown.

\subsection{Effect of the Number of Training Classes}
We examine how the number of training character classes affects trajectory recovery performance.
We change the number of classes used for training while keeping the model and training settings the same.
Using more classes also increases the number of training samples.
Figure~\ref{fig:ldtw_vs_classes} shows test-set LDTW as we increase the number of training classes, together with the corresponding number of training samples.
While the number of training samples increases almost linearly, LDTW does not improve at a constant rate.
It drops quickly up to around 500 classes and then decreases more slowly, suggesting smaller gains after the training set becomes sufficiently diverse.

One reason may be that Chinese characters are built from shared parts: many characters reuse the same radicals and local stroke patterns.
With small training classes, these common parts appear over and over, so the model can learn them quickly.
As we add more classes, many new characters are largely new combinations of parts the model has already seen, so each additional class provides less new information and the improvement slows down.

This trend is also visible in Fig.~\ref{fig:qualitative_20}, which shows qualitative results for the smallest training setting (20 classes).
The model handles simple characters well, but errors become more frequent for complex ones.
In the five worst examples, the recovered trajectories often deviate from the input image, leading to noticeable shape distortions and incorrect stroke order.
Overall, these examples suggest that limited class diversity mainly hurts shape fidelity and stroke-order estimation, both of which improve as more classes are included in training.

\subsection{Image-Conditioning Injection Strategies}
\label{sec:injection}

\begin{table}[t]
\caption{
Ablation of image-conditioning injection types.
Feature map 1/2 and 1/8 inject spatial feature maps extracted at resolutions of 1/2 and 1/8 of the input image, respectively.
Global vector uses the final image representation after a global pooling layer, which discards spatial layout information.
Multi-scale feature maps inject feature maps from multiple resolutions.
}
\label{tab:Injection}
\centering
\begin{tabular}{lccc}
\toprule
Injection type & DTW$\downarrow$ & LDTW$\downarrow$ & AIoU$\uparrow$\\
\midrule
Global vector & 191.3 & 2.629 & 0.605\\
Feature map 1/2 & 233.8 & 3.249 & 0.524\\
Feature map 1/8 & \underline{119.7} & \underline{1.625} & \underline{0.736}\\
Multi-scale feature maps & \textbf{111.1} & \textbf{1.574} & \textbf{0.745}\\
\bottomrule
\end{tabular}
\end{table}

Multi-scale spatial conditioning described in Section \ref{sec:archi} is critical for preserving glyph geometry while recovering plausible stroke order.
We ablate the image-conditioning design by varying only the form of image features provided to the denoiser: (i) a global pooled image embedding with no spatial layout (Global vector in Table~\ref{tab:Injection}), (ii) a single spatial feature map at a fixed encoder scale (Feature map 1/2 or 1/8), and (iii) the proposed multi-resolution feature maps (Multi-scale feature maps).
Here, ``1/2'' and ``1/8'' denote the spatial resolution of the feature map relative to the input image (i.e., $\frac{H}{2}\times \frac{W}{2}$ and $\frac{H}{8}\times \frac{W}{8}$), corresponding to different downsampling factors in the image encoder.\par

Table~\ref{tab:Injection} compares image-conditioning injection strategies on the seen classes test set, while keeping the remaining training and sampling settings identical to those used in Table~\ref{tab:seen}. Global vector conditioning performs poorly, indicating that compressing the image into a single global representation discards geometric evidence necessary for accurate trajectory recovery. Replacing the global vector with a spatial feature map substantially improves performance, confirming the importance of preserving 2D image structure in the conditioning pathway. Multi-scale feature-map conditioning achieves the best overall trade-off and the highest glyph fidelity, supporting our design choice of injecting image evidence at multiple resolutions.

Injecting overly fine features alone can hurt because local evidence without global layout increases ambiguity at crossings and junctions. This trend is reflected in the degraded performance of the single 1/2-resolution feature map condition, which is worse than both the 1/8-resolution and multi-scale variants. A plausible explanation is that conditioning only on fine-resolution features makes the denoiser more sensitive to local appearance variations while providing insufficient global structural context for resolving long-range stroke dependencies and stroke-order ambiguity. In contrast, multi-scale conditioning balances coarse and fine cues: coarser features provide robust global layout constraints, whereas finer features refine local geometry, leading to better shape preservation and more stable trajectory recovery.

\begin{table}[t]
\caption{Trajectory recovery performance on unseen character classes (755 classes excluded from training). Lower is better for DTW/LDTW; higher is better for AIoU.}
\label{tab:unseen}
\centering
\begin{tabular}{lccc}
\toprule
Method & DTW$\downarrow$ & LDTW$\downarrow$ & AIoU$\uparrow$\\
\midrule
Cross-VAE~\cite{Cross-VAE} & 742.6 & 7.463 & 0.337\\
PEN-Net~\cite{PEN-Net}    & \underline{272.0} & \underline{3.178} & \underline{0.594}\\
Ours             & \textbf{117.7} & \textbf{1.653} & \textbf{0.743}\\
\bottomrule
\end{tabular}
\end{table}

\begin{figure}[t]
  \centering
  \includegraphics[width=0.9\textwidth]{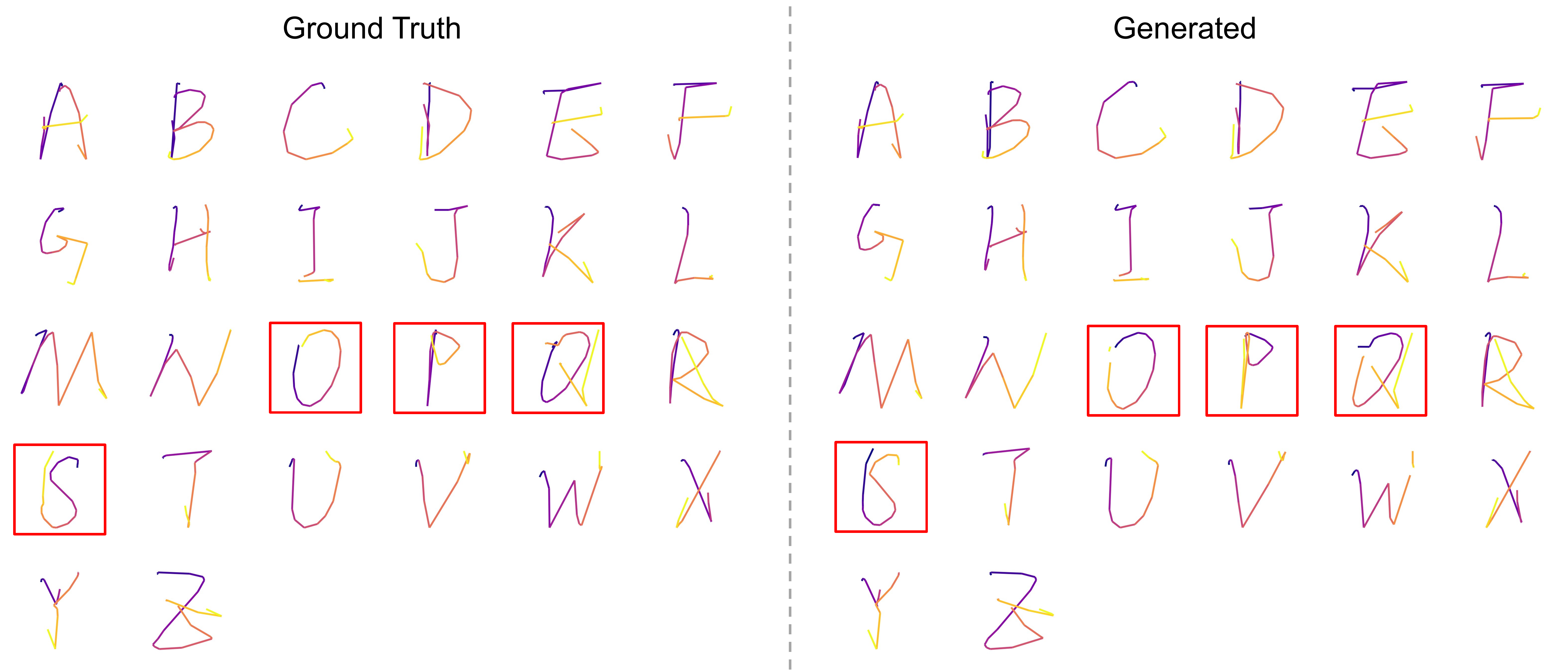}\\[-2mm]
  \caption{Qualitative results under cross-script domain shift: English characters are fed to a model trained on Chinese characters. Red boxes highlight examples where the generated trajectories differ from the recorded reference temporal order or direction.
  }
  \label{fig:domainshift}
\end{figure}

\subsection{Generalization to Unseen Character Classes}

Our method generalizes to unseen character classes and maintains strong recovery performance on held-out classes.
We evaluate on 755 character classes that are never observed during training.
Table~\ref{tab:unseen} reports the quantitative results, showing that our method still outperforms both baselines by a wide margin on the unseen-class test set.

The performance drop from the seen classes setting in Table~\ref{tab:seen} is relatively small, suggesting that the model learns transferable geometric and temporal primitives rather than memorizing class-specific trajectories.
This result is consistent with the compositional structure of Chinese characters, where many classes share reusable substructures and stroke patterns.

We evaluate cross-script generalization by applying a model trained exclusively on Chinese characters to Latin letters in CASIA-OLHWDB. As shown in Fig.~\ref{fig:domainshift}, we observe stroke-order estimation errors for a few characters (highlighted in red), which are attributable to differences in writing conventions across scripts. Because counter-clockwise stroke directions are uncommon in Chinese handwriting, the recovered trajectories for ``O'' and ``Q'' are biased toward clockwise directions.
Nevertheless, the predicted stroke order is correct for most letters, and the recovered trajectories remain consistent with the input glyph shapes without noticeable geometric deviation.

\subsection{Robustness Analyses}

We study robustness under two practical perturbations: stroke-width variation and missing strokes.
These analyses test whether the recovered trajectory remains geometrically consistent with the observable ink under degraded or out-of-domain inputs.

\begin{table}[t]
\caption{Robustness to stroke width variation (DTW/LDTW lower is better; AIoU higher is better). The top row shows example images of ``\protect\KanjiYoku'' rendered at each stroke width.}
\label{tab:width}
\centering
\begin{tabular}{lccc ccc ccc}
\toprule
& \multicolumn{3}{c}{Width=1} & \multicolumn{3}{c}{Width=2} & \multicolumn{3}{c}{Width=3}\\
& \multicolumn{3}{c}{\wimg{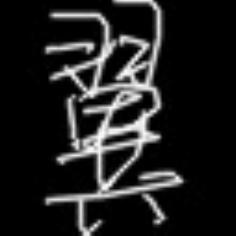}}
& \multicolumn{3}{c}{\wimg{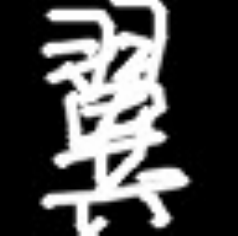}}
& \multicolumn{3}{c}{\wimg{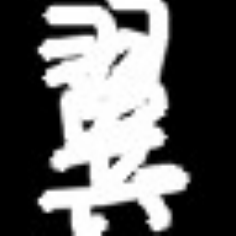}}\\[-1pt]
\cmidrule(lr){2-4}\cmidrule(lr){5-7}\cmidrule(lr){8-10}
Method & DTW$\downarrow$ & LDTW$\downarrow$ & AIoU$\uparrow$
       & DTW$\downarrow$ & LDTW$\downarrow$ & AIoU$\uparrow$
       & DTW$\downarrow$ & LDTW$\downarrow$ & AIoU$\uparrow$\\
\midrule
Cross-VAE~\cite{Cross-VAE} & 713.8 & 7.029 & 0.346 & 742.6 & 7.463 & 0.337 & 802.2 & 7.862 & 0.323\\
PEN-Net~\cite{PEN-Net}   & \underline{212.3} & \underline{2.728} & \underline{0.607} & \underline{265.0} & \underline{3.149} & \underline{0.593} & \underline{300.8} & \underline{3.403} & \underline{0.586}\\
Ours      & \textbf{111.1} & \textbf{1.574} & \textbf{0.745} & \textbf{131.1} & \textbf{1.805} & \textbf{0.741} & \textbf{156.9} & \textbf{2.121} & \textbf{0.729}\\
\bottomrule
\end{tabular}
\end{table}

\begin{figure}[t]
  \centering
  \includegraphics[width=0.8\textwidth]{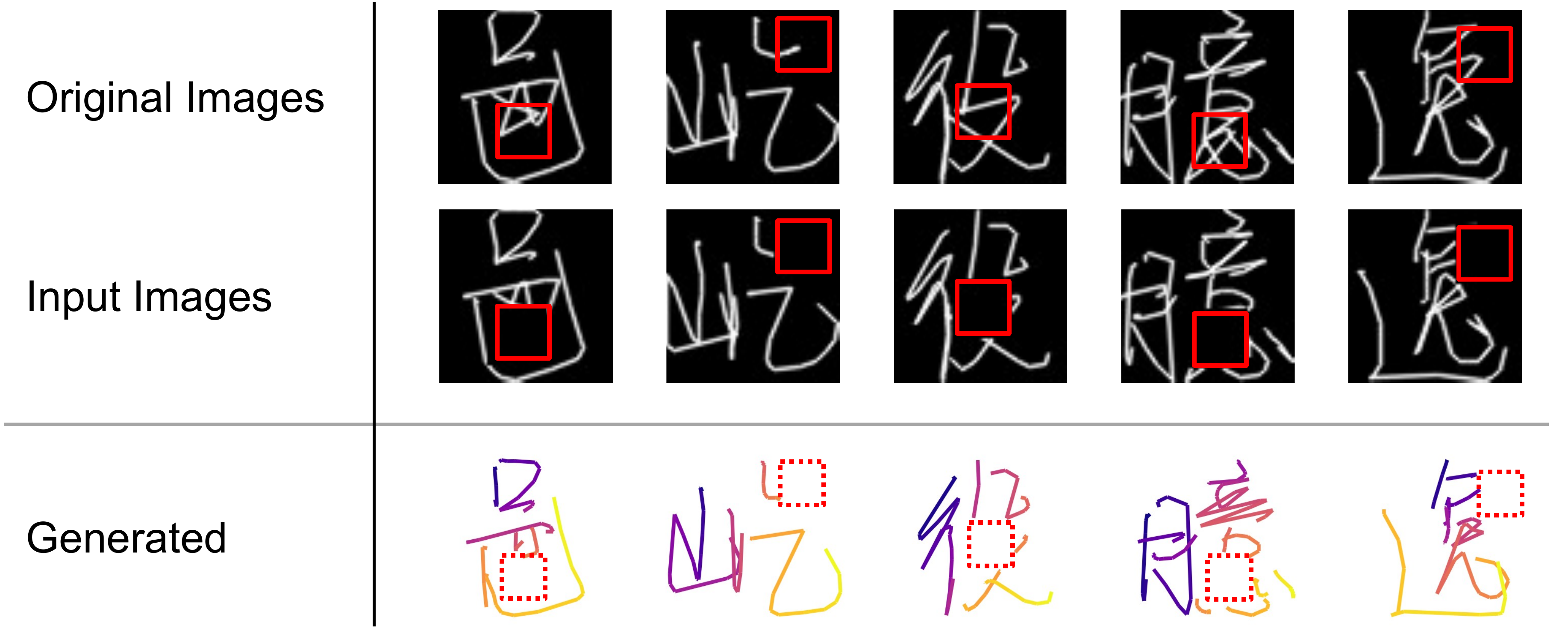}\\[-2mm]
  \caption{Recovery results of our method for character images with missing strokes (random rectangular masks). Red boxes indicate the rectangular masked regions.}
  \label{fig:missing}
\end{figure}

\paragraph{Stroke-width variation.}
We evaluate robustness to stroke-width variations because they affect local structures around junctions and crossings. Following the protocol of \cite{PEN-Net}, we train with stroke widths in $\{1,2,3\}$ pixels and evaluate separately on each condition.
Table~\ref{tab:width} shows that all methods degrade as stroke width increases, but our method remains the most accurate across all widths and exhibits only a mild drop in AIoU, indicating stable glyph preservation.

\paragraph{Missing strokes.}
To test robustness to local missing information, we train on clean images and apply random rectangular masks to the input image at inference time.
Fig.~\ref{fig:missing} shows that the model generally avoids generating spurious strokes and keeps the recovered trajectory aligned with the visible ink.
In masked regions, the model tends to be conservative: rather than aggressively filling in occluded parts, it often preserves the observed structure, and the trajectory may remain fragmented when large or critical ink regions are removed.

\section{Conclusion and Future Work}

In this paper, we address the problem of handwriting trajectory recovery using diffusion models. By formulating trajectory recovery as an image-conditioned time-series generation task, we leverage diffusion models to reconstruct temporally ordered pen trajectories that are consistent with a given offline handwriting image.
We further introduce a multi-scale conditioning technique to effectively encode image information at different resolutions. This design significantly reduces trajectory drift and improves the performance of handwriting trajectory recovery.
Experimental results on CASIA-OLHWDB demonstrate clear improvements over the baselines PEN-Net and Cross-VAE in terms of both temporal alignment (DTW/LDTW) and shape fidelity (AIoU), particularly for complex multi-stroke characters.  Furthermore, we confirm that increasing the number of training character classes leads to a significant improvement in recovery performance. These findings suggest that handwriting trajectories exhibit shared structural and dynamic regularities across character classes, and that generative models can learn underlying writing principles rather than merely memorizing character-specific stroke orders.


Several directions remain for future work. First, sampling speed could be improved by leveraging flow matching~\cite{lipman2023flow}, fewer-step samplers, or network optimization. Second, although our experiments provide preliminary evidence of cross-script transfer, extending the method to a wider range of writing systems remains an important direction. Finally, our current rendering protocol assumes uniform stroke widths, whereas real offline handwriting images may contain local stroke-width variations caused by pen pressure, writing instruments, or scanning conditions. Adapting the model to such realistic image conditions is therefore another important future direction.

\begin{credits}
\subsubsection{\ackname}
This work was supported by JSPS KAKENHI Grant Numbers JP25H01149 and JP24K20750, JST CRONOS-JPMJCS24K4, and ACT-X-JPMJAX25C.
\end{credits}

\bibliographystyle{splncs04}
\bibliography{mybibliography}

\end{document}